\documentclass[letterpaper]{article} 
\usepackage{aaai23}  
\usepackage{times}  
\usepackage{helvet}  
\usepackage{courier}  
\usepackage[hyphens]{url}  
\usepackage{graphicx} 
\usepackage{natbib}  
\usepackage{caption} 
\usepackage{algorithm}
\usepackage{algorithmic}

\usepackage{newfloat}
\usepackage{listings}
\DeclareCaptionStyle{ruled}{labelfont=normalfont,labelsep=colon,strut=off} 
\floatstyle{ruled}
\newfloat{listing}{tb}{lst}{}
\floatname{listing}{Listing}
\usepackage[utf8]{inputenc} 
\usepackage{url}            
\usepackage{booktabs}       
\usepackage{amsfonts}       
\usepackage{nicefrac}       
\usepackage{microtype}      
\usepackage{xcolor}         

\usepackage{amsmath}
\usepackage{amssymb}
\usepackage{mathtools}
\usepackage{amsthm}

\usepackage{algorithm}
\usepackage{algorithmic}
\usepackage{enumitem}
\usepackage{amsmath}
\usepackage{amsthm}
\usepackage{multirow}

\usepackage{microtype}
\usepackage{graphicx}
\usepackage{subfigure}
\usepackage{booktabs} 
\usepackage[capitalize,noabbrev]{cleveref}

\title{Learning Compact Features via In-Training Representation Alignment}
\author{
    Xin Li,
    Xiangrui Li,
    Deng Pan,
    Yao Qiang,
    Dongxiao Zhu
}
\affiliations{
    Department of Computer Science, Wayne State University

    Detroit, Michigan 48202, USA\\
    \{xinlee, xiangruili, pan.deng, yao, dzhu\}@wayne.edu

}

\usepackage{bibentry}

\begin{document}

\maketitle

\begin{abstract}
Deep neural networks (DNNs) for supervised learning can be viewed as a pipeline of the feature extractor (i.e., last hidden layer) and a linear classifier (i.e., output layer) that are trained jointly with stochastic gradient descent (SGD) on the loss function (e.g., cross-entropy). In each epoch, the true gradient of the loss function is estimated using a mini-batch sampled from the training set and model parameters are then updated with the mini-batch gradients. Although the latter provides an unbiased estimation of the former, they are subject to substantial variances derived from the size and number of sampled mini-batches, leading to noisy and jumpy updates. To stabilize such undesirable variance in estimating the true gradients, we propose In-Training Representation Alignment (ITRA) that explicitly aligns feature distributions of two different mini-batches with a matching loss in the SGD training process. We also provide a rigorous analysis of the desirable effects of the matching loss on feature representation learning: (1) extracting compact feature representation; (2) reducing over-adaption on mini-batches via an adaptive weighting mechanism; and (3) accommodating to multi-modalities. Finally, we conduct large-scale experiments on both image and text classifications to demonstrate its superior performance to the strong baselines.
\end{abstract}

\section{Introduction}

Recently, deep neural networks (DNNs) have achieved remarkable performance improvements in a wide range of challenging tasks in computer vision \cite{he2016resnet,huang2019densenet,pan2021explaining,qiang2022counterfactual}, natural language processing \cite{sutskever2014sequence,chorowski2015attention,qiang2022attcat} and healthcare informatics \cite{miotto2018deep,li2020predicting}.
For supervised learning, DNNs can be viewed as a feature extractor followed by a linear classifier on the latent feature space, which are jointly trained using stochastic gradient descent (SGD). Specifically, in each iteration of SGD, a mini-batch of $m$ samples $\{(x_i,y_i)\}_{i=1}^m$ is sampled from the training data $\{(x_i,y_i)\}_{i=1}^n (n>m)$. The gradient of loss function $L(x,\theta)$ is calculated on the mini-batch, and network parameter $\theta$ is updated via one step of gradient descent (learning rate $\alpha$):

\begin{equation}\label{eq:SGDupdate}
\begin{split}
&\frac{1}{n}\sum_{i=1}^{n}\nabla_\theta L(x_i,\theta) \approx \frac{1}{m}\sum_{i=1}^{m}\nabla_\theta L(x_i,\theta),\\
&\theta \leftarrow \theta - \alpha \cdot\frac{1}{m}\sum_{i=1}^{m}\nabla_\theta L(x_i,\theta).
\end{split}
\end{equation} 

This update in Eq.(\ref{eq:SGDupdate}) can be interpreted from two perspectives. First, from the conventional approximation perspective, the true gradient of the loss function (i.e., gradient on the entire training data) is approximated by the mini-batch gradient. As each mini-batch gradients are unbiased estimators of the true gradient of the loss function and the computation is inexpensive, large DNNs can be efficiently and effectively trained with modern computing infrastructures. Second, Eq. (\ref{eq:SGDupdate}) can also be interpreted as an exact gradient descent update on the mini-batch. In other words, SGD updates network parameters $\theta$ to achieve maximum improvement in fitting the mini-batch. As each mini-batch is often uniformly sampled from each class of the training data, such exact update inevitably introduces the undesirable variance in gradients calculation via backpropagation, resulting in the over-adaption of model parameters to that mini-batch.

\begin{figure*}[]
	\centering
	\includegraphics[scale=0.76]{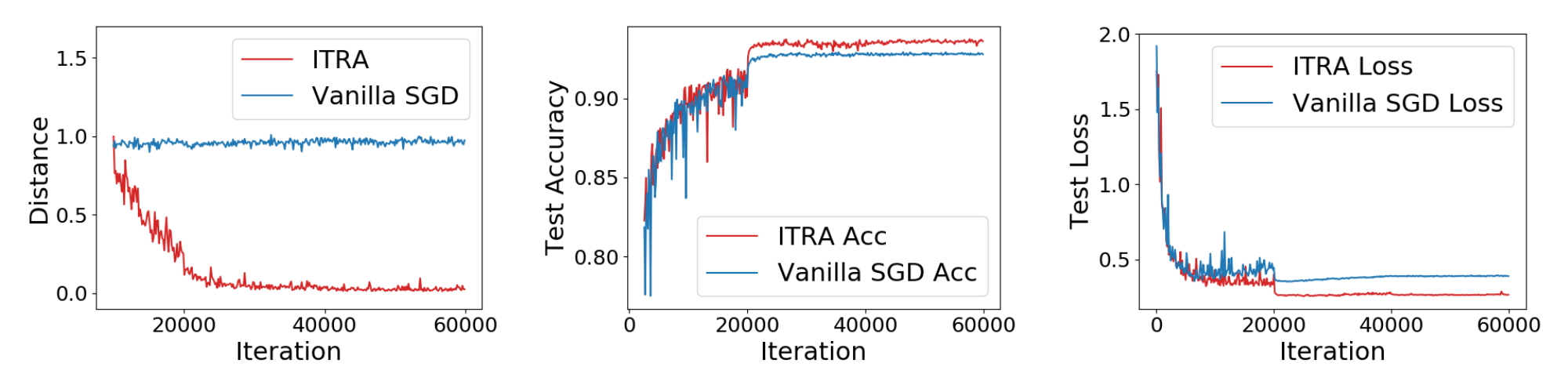}
	\caption{A comparison of ITRA and vanilla SGD training on the CIFAR10 testing data. Left: normalized distance between samples of the same class from different mini-batches used in training; middle: testing accuracy; right: testing cross-entropy loss. The model is Resnet18.}
	\label{fig:motivationExample}
	\vspace{-3mm}

\end{figure*}

A natural question then to ask is, \textit{``can we reduce the over-adaption to mini-batches?"}, to reduce the mini-batch dependence on SGD update in Eq. (\ref{eq:SGDupdate}). 
In this paper, we propose In-Training Representation Alignment (ITRA) that aims at reducing the mini-batch over-adaption by aligning feature representation of different mini-batches that is learned by the feature extractor in SGD. Our motivation for feature alignment is: \textit{if the SGD update using one mini-batch A is helpful for DNNs learning good feature representations with respect to the entire data, then for another mini-batch B, their feature representation should align well with each other}. In this way, we can reduce mini-batch over-adaption by forcing accommodation of SGD update to B and reducing dependence of the parameter update on A. Ideally, if the distribution $P(h)$ of latent feature $h$ is known as a prior, we could explicitly match the mini-batch feature $h_{\text{mb}}$ with $P(h)$ via maximum likelihood. However, in practice, $P(h)$ is not known or does not even have an analytic form. To achieve this, we utilize the maximum mean discrepancy (MMD) \cite{gretton2012kernel} from statistical hypothesis testing for the two-sample problem. MMD is differentiable that can be trained via back propagation. Moreover, we show in an analysis that the gradient of MMD enjoys several good theoretical merits.  Based on the analysis, ITRA reduces SGD update adaption to the mini-batch by implicitly strengthening the supervision signal of high-density samples via an adaptive weighting mechanism (see details in Section 4), where high-density samples are closely clustered to form modalities for each class.


To check effect of gradient update on feature representation learning, an illustrative example is presented in Figure \ref{fig:motivationExample}. The model is Resnet18 with BN layers trained with cross-entropy (CE) loss. We calculate the distance between a pair of same-class samples from two mini-batches respectively and plot the normalized distance in the left panel of Figure \ref{fig:motivationExample}, after model training stabilizes and achieves relatively good performance. We see that when model is trained only with CE loss in vanilla SGD, the distance stabilizes while the training makes progresses. This is due to that after the model capturing the classification pattern for each class, vanilla SGD adapts to mini-batch samples to achieve gain for the loss function yet does not further encourage feature alignment to learn compact feature representations. 
Hence, vanilla SGD has little effect on the compactness of feature representations. However, in ITRA, the distance between a pair of samples keeps decreasing. This implies ITRA indeed helps DNN to learn more compact feature representations by aligning different mini-batches and thus achieves higher accuracy and lower loss (Figure \ref{fig:motivationExample} middle and right panels). 


We summarize our original contributions as follows. (1) We propose a novel and general strategy ITRA for training DNNs. ITRA augments conventional SGD with regularization by forcing feature alignment of different mini-batches to reduce variance in estimating the true gradients using mini-batches. ITRA can enhance the existing regularization approaches and is compatible with a broad range of neural network architectures and loss functions. (2) We provide theoretical analysis on the desirable effects of ITRA and explains why ITRA helps reducing the over-adaption of vanilla SGD to the mini-batch. With MMD, ITRA has an adaptive weighting mechanism that can help neural networks learn more discriminative feature representations and avoid the assumption of uni-modality on data distribution. Results on benchmark datasets demonstrate that training with ITRA can significantly improve DNN performance, compared with other state-of-the-art methods.

\section{Related Work}

Modern architectures of DNNs usually have an extremely large number of model parameters, which often outnumbers the available training data. To reduce overfitting in training DNNs, regularizations are needed. Those regularization methods include classic ones such as $L_1/L_2$-norm penalties and early stopping \cite{li2018multinomial,li2018robust}. For deep learning, additional useful approaches are proposed motivated by the SGD training dynamics \cite{li2022coupling}. For example, dropout \cite{srivastava2014dropout} and its variants \cite{gao2019demystifying,ghiasi2018dropblock} achieve regularization by reducing the co-adaption of hidden neurons of DNNs. 
\cite{ioffe2015batch} proposes batch normalization (BN) to reduce the internal covariate shift caused by SGD. 
For image classification, data-augmentation types of regularization are also developed \cite{devries2017cutout,gastaldi2017shake,li2020learning,li2021improving}. Different from those approaches, our proposed ITRA is motivated by the perspective of exact gradient update for each mini-batch in SDG training, and achieves regularization by encouraging the alignment of feature representations of different mini-batches. Those methods are compatible with ITRA for training DNNs and hence can be applied in conjunction with ITRA. 

Another line of regularization are loss function based that the supervision loss is augmented with other penalties under different considerations. One example is label smoothing \cite{szegedy2016ls}, which corrupts the true label with a uniformly-distributed noise to discourage DNNs' over-confident predictions for training data. 
{\it The work that is closest to ours is Center loss} \cite{wen2016discriminative}, which reduces the intra-class variation by aligning feature of each class to its ``center". With the assumption of distribution uni-modality for each class, it explicitly encourages the feature representations clustering around its center. However, this assumption may be too strict since true data distribution is generally unknown and can be multi-model. 
On the contrary, ITRA reduces variances and encourages intra-class compactness by aligning a pair of features from two minibatches to each other, which avoids the distribution assumption and is accommodating to multi-modalities.

To match the distribution of features learned from different mini-batches, ITRA uses MMD as its learning objective. MMD \cite{gretton2007kernel,gretton2012kernel} is a probability metric for testing whether two finite sets of samples are generated from the same distribution. Using a universal kernel (i.e., Gaussian kernel), minimizing MMD encourages to match all moments of the empirical data distribution. MMD has been widely applied in many machine learning tasks. For example, \cite{li2015gmmd} and \cite{li2017mmdgan} use MMD to train unsupervised generative models by matching the generated distribution with the data distribution. Another application of MMD is for the domain adaption. To learn domain-invariant feature representations, \cite{long2015dan} uses MMD to explicitly match feature representations from different domains. There are also other probability-based distance metrics applied in domain adaption such as $\mathcal{A}$-divergence \cite{ben2007analysis} and Wasserstein distance \cite{shen2018wasserstein}. However, these metrics are {\it non-differentiable} while the differentiability of MMD enables the adaptive weighting mechanism in ITRA. Moreover, our goal is different from those applications. In ITRA, we do not seek exact distribution matching. Instead, we use class-conditional MMD as a regularization to improve SGD training.

\section{Preliminary: Maximum Mean Discrepancy}
Given two finite sets of samples $S_1 = \{x_i\}_{i=1}^n$ and $S_2 = \{y_i\}_{i=1}^m$, MMD \cite{gretton2007kernel,gretton2012kernel} is constructed to test whether $S_1$ and $S_2$ are generated from the same distribution. MMD compares the sample statistics between $S_1$ and $S_2$, and if the discrepancy is small, $S_1$ and $S_2$ are then likely to follow the same distribution.

Using the kernel trick, the empirical estimate of MMD \cite{gretton2007kernel} w.r.t. $S_1$ and $S_2$ can be rewritten as:
\begin{equation}\notag
\begin{split}
\text{MMD}(S_1, S_2) = &\big[\frac{1}{n^2}\sum_{i,j=1}^{n}\mathcal{K}(x_i,x_j) + \frac{1}{m^2}\sum_{i,j=1}^{m} \mathcal{K}(y_i,y_j) \\
&- \frac{2}{mn}\sum_{i=1}^{n}\sum_{j=1}^{m}\mathcal{K}(x_i,y_j) \big]^{1/2},
\end{split}
\end{equation}
where $\mathcal{K}(\cdot,\cdot)$ is a kernel function. \cite{gretton2007kernel} shows that if $\mathcal{K}$ is a characteristic kernel, then asymptotically MMD = 0 if and only $S_1$ and $S_2$ are generated from the same distribution. A typical choice of $\mathcal{K}$ is the Gaussian kernel with bandwidth parameter $\sigma$: $\mathcal{K}(x,y) = \exp(-\frac{||x-y||^2}{\sigma})$.
With Gaussian kernel, minimizing MMD is equivalent to matching all orders of moments of the two datasets \cite{li2015gmmd}.

\section{In-Training Representation Alignment}
\textbf{The Proposed ITRA} The idea of ITRA is to reduce the DNN over-adaption to a mini-batch if we view the SGD iteration as an exact update for that mini-batch. In terms of feature learning, we attempt to train the feature extractor to encode less mini-batch dependence into the feature representation. From the distribution point of view, the latent feature distribution of the mini-batch should approximately match with, or more loosely, should not deviate much from that of the entire data. However, aligning a mini-batch with the global statistics from entire data may not be available, we sample a pair of mini-batch to match each other to reduce the variance. It is possible to sample more mini-batches to further reduce variances but is computationally expensive.


More formally, let $f_\theta(x)$ be a convolutional neural network model for classification that is parameterized by $\theta$. It consists of a feature extractor $h = E_{\theta_e}(x)$ and a linear classifier $C_{\theta_c}(h)$ parameterized by $\theta_e$ and $\theta_c$ respectively. Namely, $f_\theta(x)= C_{\theta_c}(E_{\theta_e}(x))$ and $\theta = \{\theta_e, \theta_c\}$. Without ambiguity, we drop $\theta$ in $f, E$ and $C$ for notational simplicity. In each iteration, let $S_{(1)} = \{(x_i^{(1)},y_i^{(1)})\}_{i=1}^{m_1}$ be the mini-batch of $m_1$ samples. Then the loss function using cross-entropy (CE) on $S_{(1)}$ can be written as 
\begin{equation}\label{eq:cross-entropy}
L_{mb}(\theta) = -\frac{1}{m_1}\sum_{i=1}^{m_1} \log f_{y_i^{(1)}}(x_i^{(1)}),
\end{equation}
where $f_{y_i^{(1)}}(x_i^{(1)})$ is the predicted probability for $x_i^{(1)}$'s true label $y_i^{(1)}$. SGD performs one gradient descent step on $L_{mb}$ w.r.t. $\theta$ using Eq. (\ref{eq:SGDupdate}). To reduce $\theta$'s dependence on $S_1$ in this exact gradient descent update, we sample from the training data another mini-batch $S_{(2)}= \{(x_i^{(2)},y_i^{(2)})\}_{i=1}^{m_2}$ to match the latent feature distribution between $S_{(1)}$ and $S_{(2)}$ using MMD:
\begin{equation} \label{eq:matchlossJ}
\begin{split}
&H_{(1)} = \{h_i^{(1)}=E(x_i^{(1)}):i=1,\cdots,m_1\},\\ 
&H_{(2)} = \{h_i^{(2)}=E(x_i^{(2)}):i=1,\cdots,m_2\},\\
&\text{Match}(\theta_e; H_{(1)}, H_{(2)})  = \text{MMD}(H_{(1)},H_{(2)}).
\end{split}
\end{equation}


Our proposed ITRA modifies the conventional gradient descent step in SGD by augmenting the CE loss (Eq. (\ref{eq:cross-entropy})) with the matching loss, which justifies the name of ITRA:
\begin{equation}\label{eq:itdmj}
\theta \leftarrow \theta - \alpha \nabla_\theta \big[ L_{mb}(\theta) + \lambda\text{Match}(\theta_e;H_{(1)}, H_{(2)}) \big],
\end{equation}
where $\lambda$ is the tuning parameter controlling the contribution of the matching loss. Note that mini-batch $S_{(2)}$ is not used in the calculation of cross-entropy loss $L_{mb}(\theta)$.


\noindent\textbf{Class-conditional ITRA} For classification tasks, we could also utilize the label information and further refine the match loss as a sum of class-conditional matching loss, termed as \textbf{ITRA-c} $(k=1,\cdots,K)$:
\begin{equation}\label{eq:matchlossC}
\begin{split}
&H_{(1)}^k = \{h_i^{(1)}=E(x_i^{(1)}): y_i=k, i=1,\cdots,m_1\} \\
&H_{(2)}^k = \{h_i^{(2)}=E(x_i^{(2)}): y_i=k, i=1,\cdots,m_2\} \\
&\text{Match}_c (\theta_e; H_{(1)}, H_{(2)}) = \frac{1}{K}\sum_{k=1}^K \text{MMD}(H_{(1)}^k, H_{(2)}^k),
\end{split}
\end{equation}
where $K$ is the total number of classes and $y_i=k$ the true label of sample $x_i$. The ITRA-c update is 
\begin{equation}\label{eq:itdmc}
\theta \leftarrow \theta - \alpha \nabla_\theta \big[ L_{mb}(\theta) + \lambda\text{Match}_c(\theta_e;H_{(1)}, H_{(2)}) \big].
\end{equation}

\subsection{Analysis on ITRA}\label{sec:analysis}
\textbf{On learning compact feature representations} To further gain insight on the desirable effects of ITRA on the SGD training procedure,  we analyze the matching loss at the sample level. With the same notation in Eq. (\ref{eq:matchlossC}), the matching loss for class $k$ is 
\begin{equation}\notag
M :=\text{Match}_k =\text{MMD}(H_{(1)}^k, H_{(2)}^k).
\end{equation}
Since MMD is symmetric with respect to $H_{(1)}^k$ and $H_{(2)}^k$, without loss of generality, we consider sample $x_i^{(1)}$ with its feature representation $h_i^{(1)} = E(x_i^{(1)})$ from $H_{(1)}^k$ (but the CE loss is not symmetric and only calculated on the first mini-batch $H_{(1)}$). Then the gradient of matching loss with respect to $h_i^{(1)}$ is (superscript (1) in $x_i^{(1)}$ and $h_i^{(1)}$ are dropped.)
\begin{equation}\notag
\begin{split}
\nabla_{h_i}M = & \frac{1}{\sqrt{M}} \nabla_{h_i}\big[ \frac{1}{m_1^2}\sum_{j=1}^{m_1} \mathcal{K}(h_i, h_j^{(1)}) \\
 &- \frac{2}{m_1 m_2}\sum_{j=1}^{m_2} \mathcal{K}(h_i, h_j^{(2)}) \big].\\
\end{split}
\end{equation}

For Gaussian kernel $\mathcal{K}(x,y)$, its gradient with respect to $x$ is $\nabla_x \mathcal{K}(x,y) = -2\exp(-\frac{||x-y||^2}{\sigma})\frac{x-y}{\sigma}$. Note that $\sigma$ is data-dependent and treated as hyperparameter. Hence, it is not back propagated in the training process and in practice set as the median of sample pairwise distances \cite{gretton2012kernel,long2015dan,li2015gmmd}. By the linearity of gradient operator, we have
\begin{equation}\label{eq:matchsamplegradient}
\begin{split}
\nabla_{h_i}M 
 = &-\frac{2}{\sqrt{M}} \big[ \frac{1}{m_1^2}\sum_{j=1}^{m_1} \exp(-\frac{||h_i-h_j^{(1)}||^2}{\sigma}) \frac{h_i - h_j^{(1)}}{\sigma}\\
 &- \frac{2}{m_1 m_2}\sum_{j=1}^{m_2} \exp(-\frac{||h_i-h_j^{(2)}||^2}{\sigma}) \frac{h_i - h_j^{(2)}}{\sigma}\big].
\end{split} 
\end{equation}

We notice that for function $g_a(x)=\exp(-x^2/a)x/a$ ($a$ is some constant), $g_a(x)\rightarrow 0$ exponentially as $x\rightarrow\infty$. Hence, for fixed $\sigma$, using the triangle inequality of $L_2$ norm,
\begin{equation}\label{eq:matchsamplegradientIneq}
\begin{split}
||\nabla_{h_i}M||\leq& \frac{2}{\sqrt{M}} \big[ \frac{1}{m_1^2}\sum_{j=1}^{m_1} g_{\sigma}(||h_i-h_j^{(1)}||) \\
&+ \frac{2}{m_1 m_2}\sum_{j=1}^{m_2} g_{\sigma}(||h_i-h_j^{(2)}||) \big].
\end{split}
\end{equation}

Within the mini-batch, $\sqrt{M}$ remain as constant for all samples. From Eq. (\ref{eq:matchsamplegradientIneq}), we observe that when $x_i$ deviates significantly away from the majority of samples of the same class, i.e., noisy samples or outliers, $||h_i-h_j^{(1)}||$ and $||h_i-h_j^{(2)}||$ are large, the magnitude of its gradient in matching loss diminishes. In other words, $x_i$ will only provide signal from the supervision loss (e.g., CE loss) and its impact on matching loss is negligible. On the other hand, training ITRA with matching loss promotes the alignment of feature representations of samples that stay close in the latent feature space. From the data distribution perspective, samples deviating from the majority are likely of low-density or even outliers. Then such behavior of ITRA implies that it can help DNNs to better capture information from high density areas and reduce the distraction of ``low density" samples in learning feature representations on the data manifold.

\noindent\textbf{On reducing over-adaption to mini-batches} The analysis above shows that low-density samples only provide supervision signal in ITRA, we now analyze how ITRA reduces the over-adaption to mini-batches. It turns out that this effect is achieved by an adaptively weighted feature alignment mechanism, which \textit{implicitly} boosts the supervision signal from high-density samples and resultantly downweights relatively the contribution of low-density samples.

To understand this, we examine the full gradient of supervision loss $L$ and matching loss MMD. Note that in ITRA, the gradient of supervision loss is only calculated on one mini-batch. Without loss of generality, we consider sample $x_i$ from the first mini-batch. The full gradient of $L(x_i)$ and $M$=MMD$(x_i, H_{(2)}^k)$ with respect to $h_i$ is (using the same notation as above)
\begin{equation}\notag
\begin{split}
\nabla_{h_i}(M+L)= &\frac{4}{\sqrt{M}m_2}\sum_{j=1}^{m_2}\exp(-\frac{||h_i-h_j^{(2)}||^2}{\sigma}) \frac{h_i - h_j^{(2)}}{\sigma}\\
&+\nabla_{o_i}L \cdot\frac{\partial o_i}{\partial h_i},
\end{split}
\end{equation}
where $o_i$ is the output for $x_i$. Let $A=\sum_{j=1}^{m_2} \exp(-||h_i-h_j^{(2)}||^2/\sigma)$ and $w_j = \exp(-||h_i-h_j^{(2)}||^2/\sigma)/A$ ($\sum_{j=1}^{m_2} w_j=1$), then equivalently:
\begin{equation}\label{eq:fullgradient}
\nabla_{h_i}(M+L)= \frac{4A}{\sqrt{M}m_2\sigma} (h_i-\sum_{j=1}^{m_2}w_j h_j^{(2)})+\nabla_{o_i}L \cdot\frac{\partial o_i}{\partial h_i}.
\end{equation}
When ITRA converges and DNNs is well trained with good performance, $||\nabla_{h_i}(M+L)||\approx 0$ and $||\nabla_{o_i}L \cdot \partial o_i/\partial h_i||$ is close to zero, we have $||h_i-\sum_{j=1}^{m_2}w_j h_j^{(2)}||<\epsilon$ ($\epsilon$ is a small scalar). In other words, ITRA promotes the feature representation $h_i$ of $x_i$ to align with the weighted average $\sum_{j=1}^{m_2}w_j h_j^{(2)}(\sum_{j=1}^{m_2} w_j=1)$, where each $w_j$ is adaptively adjusted in the training process based on similarity between $h_i$ and $h_j^{(2)}$ in the latent feature space. As mini-batch samples are uniformly sampled from the training data, it is expected that on average, the majority of $\{h_j^{(2)}\}_{j=1}^{m_2}$ are from high-density area of the data distribution. For DNNs with good generalizability, DNNs must perform well for samples from those areas (as testing samples are more likely to be generated from high-density areas in the data manifold). Hence, provided that sample $x_i$ is of high-density that already provides useful supervision signal, ITRA further boosts its contribution by aligning $h_i$ with $\sum_{j=1}^{m_2}w_j h_j^{(2)}$ of other high-density samples in the 2nd mini-batch. \textit{The adaptive weight $w_j$ is critical}: if  sample $h_j^{(2)}$ is of low-density and deviates far from $x_i$, its weight $w_j$ is automatically adjusted small, having vanishing contribution in the gradient. This in turn downweights relatively the contribution of low-density samples in SGD, resulting in the reduction of over-adaption to mini-batches.

\noindent\textbf{Accommodating multi-modalities} The adaptively weighting mechanism brings another benefit: if the data distribution (for each class) is multi-modality in the latent feature space, ITRA automatically aligns $x_i$ with its corresponding modality. Specifically, without loss of generality, assume two modalities $md_1$ and $md_2$, $\{h_j^{(2)}\}$ consists of samples from $md_1$ and $md_2$ and $x_i$ is generated from $md_1$. We can rewrite $h_i-\sum_{j=1}^{m_2}w_j h_j^{(2)} = h_i-(\sum_{j\in md_1}w_j h_j^{(2)}+\sum_{j\in md_2}w_j h_j^{(2)})$. As $x_i$ is generated from $md_1$ and deviates from $md_2$, implying that $x_i$ is closer to samples from the same modality than those from the other modality. Hence, with the adaptively weighting mechanism in Eq. (\ref{eq:fullgradient}), $w_j\approx0$ $(j\in md_2)$ and $h_i-\sum_{j=1}^{m_2}w_j h_j^{(2)} \approx h_i-\sum_{j\in md_1}w_j h_j^{(2)}$. That is, align $x_i$ only with samples from the same modality. Therefore, ITRA avoids the uni-modality assumption on data distribution as in \cite{wen2016discriminative,wan2018rethinking} and justifies the advantage of nonparametric MMD for feature alignment.

\section{Experiments}

In this Section, we extensively evaluate the ITRA performance using benchmark datasets on both image classification (i.e., KMNIST \cite{clanuwat2018deep}, FMNIST \cite{xiao2017fmnist}, CIFAR10, CIFAR100 \cite{krizhevsky2009cifar}, STL10 \cite{coates2011stl10} and ImageNet \cite{deng2009imagenet}) and text classification (i.e., AG's News, Amazon Reviews, Yahoo Answers and Yelp Reviews) tasks. In our experiments, class-conditional ITRA-c is tested as it exploits implicit label information with better supervision in the training process. In addition to using vanilla SGD training as the baseline (i.e., w/o ITRA), we also compare ITRA with more widely used {\it loss-function based regularization methods} as the strong baselines for comparison: label smoothing (LSR) \cite{szegedy2016ls} and center loss (Center) \cite{wen2016discriminative}. For evaluation metrics, we report the Top-1 accuracy and CE loss value for all methods. The optimal hyperparameter value $\lambda$ for each method is also reported. Results on other tuning parameter values as well as experimental details are provided in supplementary materials.

\begin{figure*}[ht]
	\centering
	\includegraphics[scale=0.4]{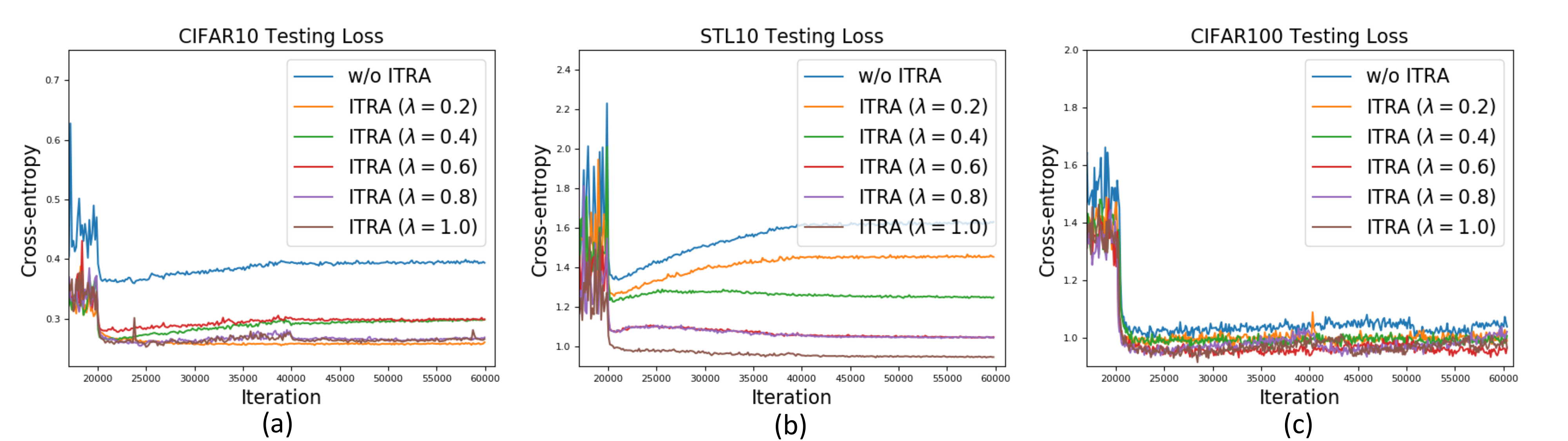}
	\caption{Testing loss of Resnet18 w.r.t. different $\lambda$ values on CIFAR10, STL10 and CIFAR100.}
	\label{fig:CIFARSTLLambdaCurve}
	\vspace{-4mm}
\end{figure*}

\noindent\textbf{Image classification} Table \ref{MNISTResults} shows the performance for KMNIST and FMNIST testing data. From the Table, we see that training with ITRA achieves better results in terms of higher accuracy and lower CE. In terms of the testing loss, ITRA has a smaller loss value compared with other methods. The testing loss with respect to different $\lambda$ values are shown in Supplementary Materials. As CE is equivalent to negative log-likelihood, smaller CE value implies that the network makes predictions on testing data with higher confidence on average. In each iteration of ITRA, there is a trade-off between the CE and matching loss. This leads to that ITRA has a regularization effect by alleviating the over-confident predictions on training data. As a result, the smaller gap between training and testing losses implies that ITRA has better generalization performance. When trained with vanilla SGD, we observe that the increasing testing loss exhibits an indication of overfitting, which is due to that FMNIST has a significant number of hard samples (e.g., those from pullover, coat and shirt classes). However, ITRA is capable of regularizing the training process hence prevents overfitting and stabilizes the testing loss as shown in the Figure \ref{fig:CIFARSTLLambdaCurve}. 
\begin{table}[t]
	\caption{Accuracy (in \%, larger is better)  and CE (smaller is better) on KMNIST and FMNIST data. The optimal performances are obtained by tuning multiple $\lambda$s according to existing literature \cite{szegedy2016ls,wen2016discriminative}.}
	\label{MNISTResults}
	\begin{center}
		\resizebox{0.4\textwidth}{!}{
			\begin{tabular}{c|ccc|ccc}
				\toprule
				&\multicolumn{3}{c|}{KMNIST}&\multicolumn{3}{c}{FMNIST}\\
				\midrule
				& $\lambda$ &  Acc $\uparrow$  & CE $\downarrow$ & $\lambda$ &  Acc $\uparrow$    &  CE $\downarrow$ \\
				\midrule
				Baseline   & -         &  95.57     & 0.183  & -         &  92.43  & 0.294   \\
				LSR & 0.1 & 95.60  & 0.181  & 0.1 & 92.47 &  0.292 \\				
				Center & 0.1 & 94.90 & 0.214  & 0.1 & 92.10  & 0.263  \\
				ITRA & 0.8 & \textbf{95.79} & \textbf{0.170}  & 0.6  &  \textbf{92.57} & \textbf{0.224}    \\
				\bottomrule
			\end{tabular} 
		}
	\end{center}
	\vspace{-3mm}
\end{table} 

\begin{table*}[th]
	\caption{Accuracy and CE loss on CIFAR10, STL10 and CIFAR100 datasets.}
	\label{tb:CIFARSTLResults}
	\begin{center}
		\resizebox{0.6\textwidth}{!}{
			\begin{tabular}{c|c|ccc|ccc|ccc}
				\toprule
				&	& \multicolumn{3}{c|}{CIFAR10} & \multicolumn{3}{c|}{STL10} & \multicolumn{3}{c}{CIFAR100}\\
				\midrule
				
				&	& $\lambda$ &  Acc $\uparrow$  & CE $\downarrow$ & $\lambda$ &  Acc $\uparrow$   & CE $\downarrow$ & $\lambda$ &  Acc $\uparrow$  & CE  $\downarrow$\\
				\midrule
				
				\multirow{4}{*}{Resnet18} &	Baseline   & -  &  92.99 & 0.40  & -  &  70.88  & 1.63  &-  & 74.19  &  1.05    \\
				
				&LSR   &  0.1  &  92.73   & 0.42  & 0.1  &  71.08  &  1.55 &0.1 & 74.21   & 1.04  \\
				
				&Center   & 0.1  &  92.30    & 0.35 &   0.1  &  70.97   & 1.10 &0.05 &  73.98 & 0.98 \\
				
				& ITRA  & 0.8  &  \textbf{93.70}  & \textbf{0.27} &  0.6 &  \textbf{72.78} & \textbf{1.05} &0.6   &  \textbf{74.88}    & \textbf{0.97} \\
				\midrule
				
				\multirow{4}{*}{VGG13} &	Baseline   & -   &  92.49    & 0.47    & -   &  74.40   & 1.55 &- & 71.72 & 1.46 \\
				
				& LSR   &  0.1  &  92.53  & 0.46 &  0.1    &  74.50  & 1.51 &0.1 & 71.75 & 1.43 \\
				
				& Center   & 0.05  &  92.11  & 0.38  & 0.05  &  74.04 & 1.16 &0.05 & 71.65 & 1.31 \\
				
				& ITRA  & 0.8   &  \textbf{92.72}  & \textbf{0.33} & 0.8  &  \textbf{75.80} & \textbf{0.93} & 0.6  & \textbf{72.55} & \textbf{1.22} \\
				\midrule
				
				\multirow{4}{*}{MobileV2} &	Baseline   & -   &  88.55    & 0.62    &    - & 59.09  & 2.14 &- & 66.42 & 1.57\\
				
				& LSR   &  0.1         &  88.77  & 0.61    &  0.1  & 59.01  &  2.12 &0.1 & 66.60   & 1.55\\
				
				& Center   & 0.1         &  88.81  &   0.53    & 0.1  & 58.24  &  \textbf{1.46} &0.05 & 66.39  & 1.51 \\
				
				& ITRA  & 1.0       &  \textbf{89.37} & \textbf{0.43}    &  0.6 &  \textbf{62.02} & 1.60 &0.6 &  \textbf{67.23}  & \textbf{1.49 }  \\
				\bottomrule
			\end{tabular}	

			}
	\end{center}
	
\end{table*} 

\begin{table}[]
	\caption{Accuracy and CE loss of Resnet-101 on CIFAR-100 and Resnet-18 on ImageNet.}
	\label{tb:large}
	\begin{center}
	\resizebox{0.48\textwidth}{!}{
\begin{tabular}{@{}c|ccc|ccc@{}}
\toprule
                     & \multicolumn{3}{c|}{CIFAR-100 (Resnet-101)} & \multicolumn{3}{c}{ImageNet (Resnet-18)} \\ \midrule
                     & Batch Size       & Acc         & CE        & Batch Size      & Acc        & CE        \\ \midrule
Baseline             & 200              & 75.85       & 1.04      & 256             & 50.01      & 2.24      \\ \midrule
Center   ($\lambda=0.05$)           & 200              & 75.23       &   1.05    & 256             &  51.83     &  2.21     \\ \midrule
\multirow{4}{*}{ITRA ($\lambda=0.6$)}  & 50               & 75.97       & 1.01      & 64              & 51.32      & 2.19      \\
 & 100              & \textbf{77.35}       & 1.01      & 128             & 51.69      & 2.17     \\ 
  & 200              &76.85        & \textbf{0.94}       & 256             & \textbf{53.13}      & \textbf{2.07}      \\\bottomrule
\end{tabular}
}
\end{center}
\vspace{-3mm}
\end{table}

Additionally, in Table \ref{tb:CIFARSTLResults}, we present the performance of Resnet18, VGG13 and MobilenetV2 on CIFAR10, STL10 and CIFAR100. From the Table, we see that ITRA achieves the best performance compared among all the four methods. Especially for the relatively more challenging (lower accuracy) STL10 data set, ITRA outperforms the baseline with a significant margin, i.e., Resnet $1.9\%$, VGG13 $1.4\%$ and MobilenetV2 $2.9\%$. In terms of CE loss, all methods have similar training losses that are close to zero. However, ITRA and Center have significantly better testing loss than other the baseline and LSR. A closer gap between training and testing losses indicates a better generalization of the DNN models enabled by regularization capability of ITRA.


\begin{table*}[th]
	\caption{Accuracy and CE loss on text classification task.}
	\label{tb:TextResults}
	\begin{center} 
	\resizebox{0.75\textwidth}{!}{
			\begin{tabular}{c|c|ccc|ccc|ccc|ccc}
				\toprule
				&	& \multicolumn{3}{c|}{AG's News} & \multicolumn{3}{c|}{Amazon Full}  &\multicolumn{3}{c|}{Yahoo Answers} &\multicolumn{3}{c}{Yelp Full} \\
				\midrule
		
				&	& $\lambda$ &  Acc $\uparrow$  & CE $\downarrow$ & $\lambda$ &  Acc $\uparrow$   & CE $\downarrow$ & $\lambda$ &  Acc $\uparrow$  & CE $\downarrow$ & $\lambda$ &  Acc $\uparrow$  & CE $\downarrow$ \\
				\midrule
				
				\multirow{3}{*}{Bert-base} &	Baseline   & -  &  92.10 & 0.42  & -  &  58.53  & 1.66  & -  &  66.60 & 1.36  & -  &  61.12 & 1.52   \\
				
				&Center   & 0.1  &  91.70    & 0.45 &   0.1  &  59.38   & \textbf{1.23}  & 0.1  &  66.95    & 1.25 & 0.1  &  60.58    & 1.17 \\
				
				& ITRA  & 0.5  &  \textbf{92.45}  & \textbf{0.29} &  0.2 &  \textbf{60.08} & 1.29 & 0.6  &  \textbf{67.47}  & \textbf{1.25} & 0.4  &  \textbf{61.68}  & \textbf{1.09} \\
				
				\midrule
				\multirow{3}{*}{DistillBert} &	Baseline   & -   &  92.13    & 0.38    & -   &  58.10   & 1.38 & -   &  66.50    & 1.28 & -   &  60.02    & 1.32  \\
				
				& Center   & 0.1  &  91.33  & 0.47  & 0.1  &  57.60 & 1.21 & 0.1  &  66.57  & 1.26 & 0.1  &  59.58  & \textbf{1.14}  \\
				
				& ITRA  & 0.2   &  \textbf{92.27}  & \textbf{0.26} & 0.4  &  \textbf{58.30} & \textbf{1.20} & 0.1   &  \textbf{66.65}  & \textbf{1.17} & 0.1   &  \textbf{60.30}  & 1.17 \\
	
				\midrule
				
				\multirow{3}{*}{XLNet} &	Baseline   & -   &  91.43    & 0.40    &    - & 60.65  & 1.34  & -   &  66.90    & 1.29  & -   &  62.78    & 1.21 \\

				& Center   & 0.1         &  90.95  &   0.45    & 0.1  & 59.88  &  \textbf{1.18} & 0.1  & 66.93  &  1.28 & 0.1  & 62.82  &  1.10 \\
				
				& ITRA  & 0.8       &  \textbf{91.85} & \textbf{0.27}    &  0.5 &  \textbf{60.90} & 1.23  &  0.5 &  \textbf{67.05} & \textbf{1.15} &  1.0 &  \textbf{63.32} & \textbf{1.08} \\
				\bottomrule
			\end{tabular}
		}
	\end{center}

\end{table*}

\noindent\textbf{Larger-scale experiment on image classification} Table \ref{tb:large} shows results on the large-scale ImageNet dataset and the deeper Resnet-101 network architecture. Note that compared with performance of Resnet-18 in Table \ref{tb:CIFARSTLResults}, the deeper Resnet-101 indeed demonstrates a better performance over the CIFAR100 dataset. For both larger dataset and deeper network, ITRA consistently achieves better accuracy and lower CE value than other methods. Markedly, for the ImageNet dataset, ITRA improves the accuracy by $5.0\% (3.12/62.92)$ and CE value by $9.7\% (0.15/1.55)$  over the standard baseline. When compared with the strong baseline Center loss, ITRA also improves the accuracy by $2.1\% (1.36/64.68)$ and CE value by $4.8\% (0.07/1.47)$. The training time of ResNet-101 (CIFAR-100) and ResNet-50 (ImageNet) using ITRA are $7.5\%$ and $3.9\%$ more than baseline on an RTX 3090 GPU. Despite the moderate increase in training time for large-scale experiments due to the extra computation incurred by sampling additional mini-batch, it demonstrates a reasonable trade-off between the increase in computational cost and gaining attractive analytic properties of ITRA.

\noindent\textbf{Large-scale experiment on text classification} ITRA performance is also evaluated on large-scale text classification experiments. We use different loss functions for fine-tuning the pre-trained Bert-base, DistillBert and XLNet models from Huggingface transformers library \cite{wolf2019huggingface}. Table~\ref{tb:TextResults} shows that the models fine-tuned with ITRA achieve a better performance in terms of accuracy and CE value on most datasets. Specifically, for Bert-base, DistillBert, XLNet models, ITRA achieves an average accuracy improvement of $1.3\%$, $0.3\%$, $0.5\%$, respectively, and an average CE value improvement of $22.4\%$, $16.1\%$, and $15.6\%$, respectively. It is worth noting that the Center loss also reduces CE value occasionally in the experiments, its accuracy performance is nevertheless compromised. The potential reason behind this phenomenon could be the multi-modality of natural language. The Center loss's uni-modality assumption helps model to minimize the distances within the class (hence CE), but can therefore lead to sub-optimal feature learning for hard samples near the boundary between modals if the class conditional distribution is indeed multi-modal.

\begin{figure*}[ht]
	\centering
	\includegraphics[scale=0.75]{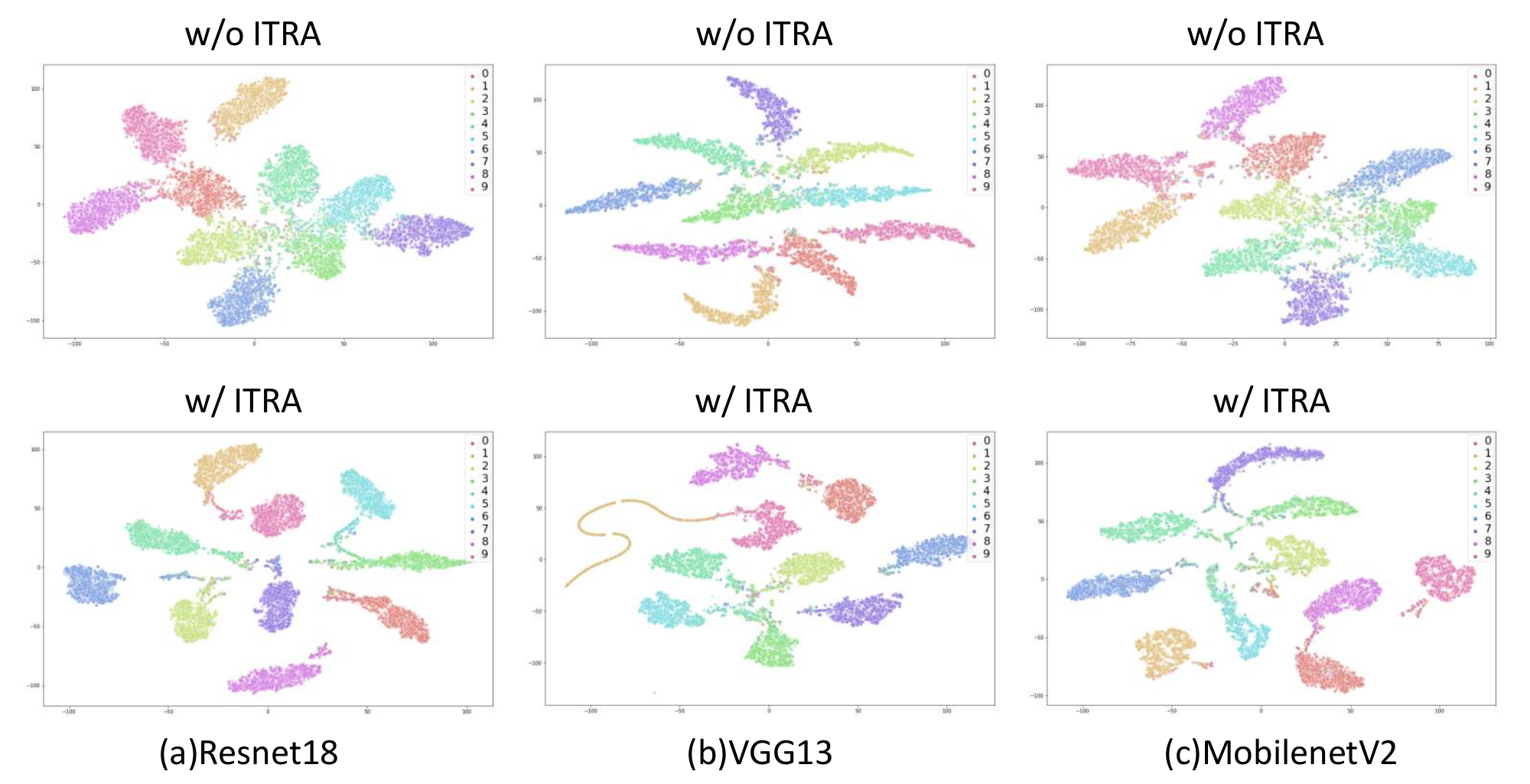}
	\caption{T-SNE plot for CIFAR10 testing data. Networks are trained with $\lambda$ that achieves best accuracy in Table \ref{tb:CIFARSTLResults}.}
	\label{fig:CIFARTSNE}
	\vspace{-3mm}
\end{figure*}

\noindent\textbf{Comparing with Center loss} As discussed in Section 2, center loss \cite{wen2016discriminative} is the closest work to ours. It effectively characterizes the intra-class variations by aligning features of each to its ``center" which is designed to reduce variance in feature learning and results in compact feature representations. Different from ITRA which aligns a pair of features from two minibatches to each other, Center loss explicitly assumes uni-modality of data distribution at feature level for each class, which may be valid in face recognition task where the Center loss is initially proposed for, but can be too stringent in classification task as class-conditional density can be multi-modal. On the contrary, ITRA is capable of accommodating the multi-modalities supported by a rigorous analysis in Section 4.1. In Figure \ref{fig:CIFARTSNE}, the model trained with ITRA effectively captures the ``typical pattern" of each class at feature level and misses some hard samples to improve generalizability. The results on both image and text classifications concur with our analysis: as shown in the Tables \ref{MNISTResults}, \ref{tb:CIFARSTLResults} and \ref{tb:TextResults}, although Center loss can occasionally reduce the CE value, it is still outperformed by our method due to its strong assumption. 

\noindent\textbf{Hyperparameter $\lambda$ and batch size} Here we also investigate the influence of hyperparameter $\lambda$ and batch size of ITRA. As shown in the Tables \ref{MNISTResults}\&\ref{tb:CIFARSTLResults}, when $\lambda$ is set with a relatively large value of 0.8 or 1, ITRA can outperform other methods in terms of both accuracy and testing loss, which is due to that larger $\lambda$s incorporate stronger implicit supervision information as mini-batches from the same class are matched. We also plot the CE loss for different $\lambda$s in Figure \ref{fig:CIFARSTLLambdaCurve} w.r.t. Resnet. Comparing with baseline, we see that training with ITRA results in significant gain in CE, regardless of network architecture. Looking at Figure \ref{fig:CIFARSTLLambdaCurve} (b) in more detail, when trained with the baseline, the testing loss shows an increasing trend as a sign of overfitting while ITRA can alleviate this trend as $\lambda$ increases. For batch size, Table \ref{tb:large} demonstrates that the increase of batch size indeed helps the reduction of variation in feature learning (lower CE loss), however, it usually requires advanced large-scale GPU clusters. Table \ref{tb:large} illustrates that ITRA achieves a better performance even with $1/4$ batch size compared to the baseline and thus avoid this hardware restriction. 

\noindent\textbf{Learning compact feature representations} From the geometric perspective, samples from the same class should stay close (i.e., intra-class compactness) and those from different classes are expected to stay far apart (i.e., inter-class separability) in the feature space (so that $f_k$ output by softmax is large). We visualize the distribution of CIFAR10 testing samples with T-SNE \cite{maaten2008tsne} in Figure \ref{fig:CIFARTSNE}. From the figure, we have the following observations: (1) ITRA learns feature representation that is much tighter with clearer inter-class margin than that learned by vanilla SGD training. (2) The data distribution in the latent space learned by ITRA exhibits a consistent pattern that for each class, the majority of testing samples are closely clustered to form a data manifold, while a small subset of samples deviate from the majority. This phenomenon concurs with our analysis that the matching loss provides diminishing gradient signals for ``low-density" samples while encourages the closeness of ``high-density" samples. Hence, ITRA can effectively capture the ``typical pattern" of each class but can miss some hard samples that overlap with other classes. This explains why ITRA achieves impressive improvement in CE value but not as much in accuracy. Overall, ITRA still outperforms vanilla SGD training and can be used as a promising training prototype that enjoys theoretical merits as shown in the analysis for matching loss.

\section{Conclusion}
In this paper, we propose a new training strategy, ITRA, as a loss function based regularization approach that can be embedded in the standard SGD training procedure. ITRA augments vanilla SGD with a matching loss that uses MMD as the objective function. We show that ITRA enjoys three theoretical merits that can help DNN learn compact feature representations without assuming uni-modality on the feature distribution. Experimental results demonstrate its excellent performance on both image and text classification tasks, as well as its impressive feature learning capacity. We outline two possible directions for future studies. The first is to improve ITRA that can learn hard sample more effectively. The second is the ITRA application in learning from poisoned datasets as ITRA is able to capture the high density areas (i.e., modalities) for each class where poisoned samples deviates far from those areas (e.g., erroneously labeled samples from other classes).

\newpage

\section{Acknowledgements}
This work is supported by the National Science Foundation under grant IIS-2211897.

\section{Ethical Impact}
This paper proposes ITRA to improve the performance of feature representation learning for training DNNs. As a general training strategy for supervised classification problems, ITRA can be used as a drop-in replacement in wherever the vanilla SGD is used. Most modern deep learning model training utilizes SGD as the standard  optimization algorithm where researchers have already proposed various alternatives and/or enhancements to  overcome its intrinsic limitations; the over adaption to mini-batch is more pronounced. The proposed in-training regularization via aligning representations enables learning more compact features thus to improve the generalizability of the predictive models. With our and many other effective feature representation learning approaches, manual feature engineering requiring profound domain knowledge and expertise will eventually phase out. As such, our research has positive impacts to a broad range of machine learning and artificial intelligence domains where domain adaption and generalization become the primary concern. For example, in medical imaging based diagnosis, leveraging ITRA on a smaller labeled and heterogeneous training set is expected to demonstrate a competitive and consistent performance to other medical imaging data sets.   

\bibliography{aaai23.bib}

\clearpage

\section*{Supplementary Materials}

\begin{table}[th]
	\caption{Text classification tasks datasets statistics.}
	\label{tb:datasets statistics}
	\begin{center}
		\begin{small}
			\begin{tabular}{c|c|c|c}
				\toprule
				Dataset & Classes  & Train Samples   & Test Samples     \\
				\midrule
				AG's News   & 4 & 120,000 & 76,00  \\
				\midrule
				Amazon Full   & 5 & 650,000 & 50,000 \\
				\midrule
				Amazon Polarity  & 2 & 560,000 & 38,000  \\
				\midrule
				Yahoo Answers    & 10 & 1,400,000 & 60,000  \\
				\midrule
				Yelp Full & 5 & 3,600,000 & 650,000 \\
				\midrule
				Yelp Polarity & 2 & 3,000,000 & 400,000 \\
				\bottomrule
			\end{tabular}
		\end{small}
	\end{center}
\end{table}
\medskip

\noindent\textbf{Comparison} In the experiments, in addition to the vanilla SGD training as the baseline (i.e., w/o ITRA), we also compare ITRA with other loss-function based regularization methods.  All DNNs in our experiments include BN layers as part of the model architectures. Label smoothing \cite{szegedy2016ls} (LSR) is a target-based regularization that the Dirac distribution for ground truth label is replaced with a mixture of Dirac distribution and uniform distribution; center loss \cite{wen2016discriminative} (Center) encourages interclass compactness of feature representations
which augments the cross-entropy loss with the maximum likelihood \cite{wan2018rethinking} assuming that each class follows Gaussian distribution in latent feature space. 

\medskip

\noindent\textbf{Implementation Details} Through all experiments, the optimization algorithm is the standard stochastic gradient descent with momentum and the loss function is cross-entropy (CE) loss. In ITRA-c, CE loss is further combined with the matching loss Eq. (\ref{eq:itdmc}) in each iteration. 

For the initial experiments on QMNIST, KMNIST and FMNIST, the CNN architecture of 2 convolutional layers is Conv(C20K5S1)-MP(K2S2)-Conv(C50K5S1)-MP(K2S2)-100-10, where CxKySz for a convolutional layer means x convolutional kernels with kernel size y, stride z; MP represents max-pooling. To train the network, we use the SGD algorithm with a momentum of 0.5. The number of epochs is 50. Learning rate is 0.01 and multiplied by 0.2 every 20 epochs. Batch size is set to 150. We don't use $L_2$ regularization. In ITRA, the tuning parameter $\lambda$ is set to 1.

 \begin{table}[th]
	\caption{Test accuracy w.r.t. different $\lambda$s.}
	\label{tb:LSRlbd}
	\begin{center}
		\begin{small}
			\begin{tabular}{c|cccc}
				\toprule
				&       \multicolumn{4}{c}{LSR}       \\
				\midrule
				$\lambda$& 0.05  & 0.1   & 0.15  & 0.2    \\
				\midrule
				KMNIST   & 95.51 & \textbf{95.60} & 95.18 & 95.02  \\
				\midrule
				FMNIST   & 92.39 & \textbf{92.47} & 92.07 & 91.83  \\
				\midrule
				CIFAR10  & 92.68 & \textbf{92.73} & 92.33 & 92.06  \\
				\midrule
				STL10    & 71.05 & \textbf{71.08} & 70.95 & 70.94  \\
				\midrule
				CIFAR100 & 74.19 & \textbf{74.21} & 73.91 & 73.72 \\
				\bottomrule
			\end{tabular}
 \text{ }\\
\medskip
 \text{ }\\
		\begin{tabular}{c|cccc}
			\toprule
			&         \multicolumn{4}{c}{Center}       \\
			\midrule
			$\lambda$ & 0.05  & 0.1    & 0.15  & 0.2   \\
			\midrule
			KMNIST    & 94.78 & \textbf{94.91}   & 94.62 & 94.14 \\
			\midrule
			FMNIST   & 91.99 & \textbf{92.10}  & 91.63 & 91.27 \\
			\midrule
			CIFAR10  & 92.16 & \textbf{92.30}   & 91.97 & 91.63 \\
			\midrule
			STL10    & 70.82 & \textbf{70.97}  & 70.23 & 69.94 \\
			\midrule
			CIFAR100 & \textbf{73.98} & 73.84  & 73.26 & 72.86\\
			\bottomrule
		\end{tabular}
\text{ }\\
\medskip
\text{ }\\
			\begin{tabular}{c|ccccc}
				\toprule
				&        \multicolumn{5}{c}{ITRA}      \\
				\midrule
				$\lambda$& 0.2   & 0.4   & 0.6   & 0.8   & 1     \\
				\midrule
				KMNIST   & 95.68 & 95.59 & 95.75 & \textbf{95.79} & 95.7  \\
				\midrule
				FMNIST   & 92.42 & 92.44 & \textbf{92.57} & 92.51 & 92.45 \\
				\midrule
				CIFAR10  & 93.52 & 92.91 & 92.93 & \textbf{93.71} & 92.98 \\
				\midrule
				STL10    & 71.64 & 72.78 & \textbf{72.88} & 71.29 & 72.53 \\
				\midrule
				CIFAR100 & 74.16 & 74.15 & \textbf{74.88} & 74.43 & 74.35\\
				\bottomrule
			\end{tabular}
		\end{small}
	\end{center}
\end{table}

For KMNIST and FMNIST in ``Experiments" section, we build a 5-layer convolutional neural network (CNN) with batch normalization applied. The CNN architecture  is Conv(C32K3S1) - BN - Conv(C64K3S1) - BN - Conv(C128K3S1) - MP(K2S2) - Conv(C256K3S1) - BN - Conv(C512K3S1) - MP(K8S1) - 512 - 10, where BN represents batch-normalization. Momentum is set to 0.5, batch size 150, number of epochs 50, initial learning rate 0.01 and multiplied by 0.2 at 20th and 40th epoch. No data augmentation is applied. For CIFAR10 and STL10, we use publicly available implementation of VGG13 \cite{simonyan2014very}, Resnet18 \cite{he2016resnet} and MobilenetV2 \cite{howard2017mobilenet}. All models are trained with 150 epochs, SGD momentum is set to 0.5, initial learning rate is 0.5 and multiplied by 0.1 every 50 epochs, batch size 150. We resize STL10 to 32 $\times$ 32. For colored image datasets, we use random crop and horizontal flip for data augmentation. For CIFAR100 with more classes, we use weight decay of $5\times10^{-4}$ and a mini-batch size of 200. For large ImageNet dataset, we trained a plain Resnet18 for 45 epochs with initial learning rate of 0.2 and multiplied by 0.1 every 15 epochs. For text classification tasks, we fine-tune the pre-trained Bert-base, DistillBert and XLNet models for 4 epochs with learning rate as $5\times10^{-5}$ and weight decay as $1\times10^{-5}$ for all comparison methods on all the datasets. 

In all experiments, networks are trained with each method under the same setting (learning rate, batch size et al.). For the bandwidth parameter in Gaussian kernels, we follow the practice in \cite{gretton2007kernel,gretton2012kernel,long2015dan} that takes the heuristic of setting $\sigma$ as the median squared distance $\sigma_{\text{Med}}$ between two samples and use a mixture of Guassian kernels with bandwidth set as a multiple of $\sigma_{\text{Med}}$. In our experiments, we use 5 kernels $k_{\text{mix}}(x,y) = \frac{1}{5}\sum_{i=1}^{5}k_{\sigma_i}(x,y)$ with $\{\sigma_i = 2^i\sigma_{\text{Med}}: i=0,\cdots,4\}$. To select the tuning parameter $\lambda$ in each method, we split out 5,000 images from the training data as validation dataset. For $\lambda$ in ITRA, we test \{0.2, 0.4, 0.6, 0.8, 1\} for checking ITRA's sensitivity to it. Note that when $\lambda=0$, ITRA is equivalent to vanilla SGD training. In LSR, we select the tuning parameter from $\{0.2,0.15,0.1,0.05\}$; for center loss, $\{0.2,0.15,0.10,0.05\}$ are tested. We utilize the standard train/test split given in the benchmark datasets and train all models once on the training data and performances are evaluated on the testing data. Performance w.r.t. different $\lambda$ values for each method is shown in Table \ref{tb:LSRlbd}.

\begin{figure}[t]
	\centering
	\includegraphics[scale=0.48]{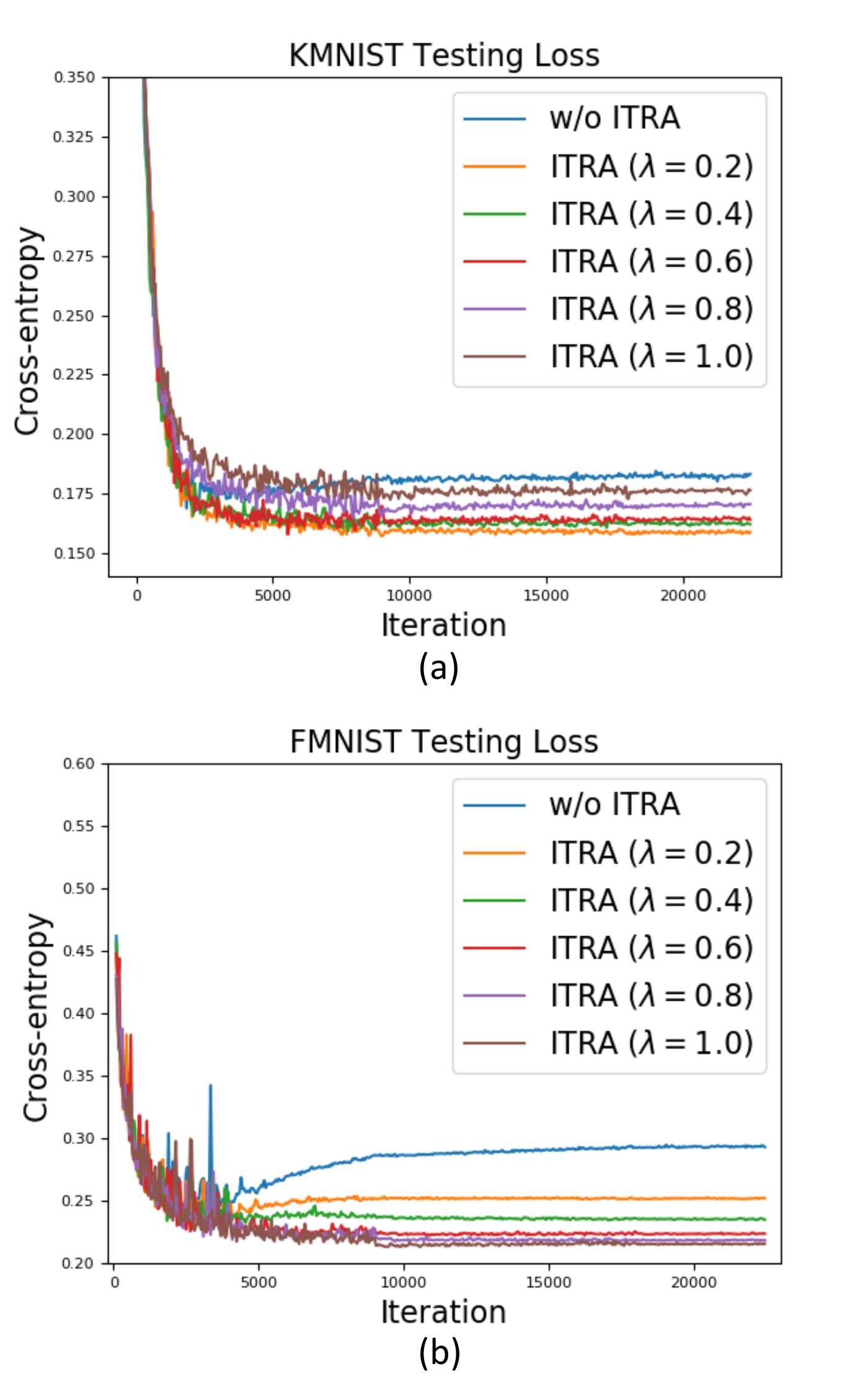}
	\caption{Testing loss w.r.t. different $\lambda$ values on KMNIST and FMNIST}
	\label{fig:KMNISTFMNISTLambdaCurve}
\end{figure}

 \noindent\textbf{KMNIST and FMNIST loss curve} We plot the testing loss with respect to different $\lambda$ values for KMNIST and FMNIST in Figure \ref{fig:KMNISTFMNISTLambdaCurve} (a) and (b) respectively. We can observe that for all $\lambda$ values, ITRA always achieves better testing loss than the baseline (w/o ITRA). Particularly for FMNIST, the baseline exhibits overfitting with an increasing trend of testing loss while ITRA provides regularization that stabilizes the testing loss.

\end{document}